\documentclass[conference]{IEEEtran}
\IEEEoverridecommandlockouts
\usepackage{cite}
\usepackage{amsmath,amssymb,amsfonts}
\usepackage{algorithmic}
\usepackage{graphicx}
\usepackage{textcomp}
\usepackage{booktabs}
\usepackage{multirow}
\usepackage{url}
\usepackage{comment}
\usepackage{flushend}
\def\BibTeX{{\rm B\kern-.05em{\sc i\kern-.025em b}\kern-.08em
    T\kern-.1667em\lower.7ex\hbox{E}\kern-.125emX}}
\begin{document}

\title{PatchBMI-Net: Lightweight Facial Patch-based Ensemble for BMI Prediction}

\author{
\IEEEauthorblockN{Parshuram N. Aarotale}
\IEEEauthorblockA{\textit{Dept. of Biomedical Engineering} \\
\textit{Wichita State University} \\
 \textit{Kansas, USA}\\
pnaarotale@shockers.wichita.edu}
\and
\IEEEauthorblockN{Twyla Hill}
\IEEEauthorblockA{\textit{Dept. of Sociology} \\
\textit{Wichita State University} \\
\textit{Kansas, USA}\\
twyla.hill@wichita.edu}
\and
\IEEEauthorblockN{Ajita Rattani}
\IEEEauthorblockA{\textit{Dept. of Computer Science and Engineering} \\
\textit{Uni. of North Texas at Denton} \\
\textit{Texas, USA}\\
ajita.rattani@unt.edu}}
\maketitle

\begin{abstract}
Due to an alarming trend related to obesity affecting $93.3$ million adults in the United States alone, body mass index~(BMI) and body weight have drawn significant interest in various health monitoring applications. 
Consequently, several studies have proposed self-diagnostic facial image-based BMI prediction methods for healthy weight monitoring. These methods have mostly used convolutional neural network (CNN) based regression baselines, such as VGG19, ResNet50, and EfficientNetB0, for BMI prediction from facial images. However, the high computational requirement of these heavy-weight CNN models limits their deployment to resource-constrained mobile devices, thus deterring weight monitoring using smartphones. This paper aims to develop a lightweight facial patch-based ensemble (PatchBMI-Net) for BMI prediction to facilitate the deployment and weight monitoring using smartphones. 
Extensive experiments on BMI-annotated facial image datasets suggest that our proposed PatchBMI-Net model can obtain Mean Absolute Error~(MAE) in the range [$3.58$, $6.51$] with a size of about $3.3$ million parameters.
On cross-comparison with heavyweight models, such as ResNet-50 and Xception, trained for BMI prediction from facial images, our proposed PatchBMI-Net obtains equivalent MAE along with the model size reduction of about $5.4\times$ and the average inference time reduction of about $3\times$ when deployed on Apple-$14$ smartphone. Thus, demonstrating performance efficiency as well as low latency for on-device deployment and weight monitoring using smartphone applications.

\end{abstract}

\begin{IEEEkeywords}
Body Mass Index, Deep Learning, Visual Attributes, Facial Images, Convolutional Neural Networks, On-device AI
\end{IEEEkeywords}

%
\section{Introduction}
\label{sec:intro}
Any visual or contextual information that can be automatically gleaned from images is known as a describable visual attribute~\cite{6248021,selfie}. Such attributes can be broadly categorized as demographic (such as age and gender), anthropometric (facial geometry), medical (such as BMI and various health conditions), material (such as spectacles and scarves), and behavioral (such as gait)~\cite{bio_iet,8272766,jain04}. 
These attributes have drawn significant interest in various applications such as surveillance, forensics, human-computer interaction, indexing, targeted advertisement systems, and health-care~\cite{8272766}. 

Obesity is one of the biggest drivers of preventable chronic diseases and healthcare costs in the United States\footnote{https://www.cdc.gov/obesity/}. Chronic conditions related to obesity include heart disease, stroke, type $2$ diabetes, and some cancers, which are the leading causes of preventable death. Severe obesity costs the United States approximately $69$ billion overall, with almost $8$ billion a year being paid for via state Medicaid programs~\cite{Wang}. 

Recently, body weight and BMI have attracted a lot of attention in applications involving health monitoring and weight loss~\cite{Wen13,8546159,Kocabey} as a result of the alarming trend in obesity. 
BMI is defined as (Body Mass in Kilograms)/(Body Height in Meters)$^2$. A BMI of $25.0$ to $30.0$ and $30$ or higher falls within the overweight and obese range, respectively. 


To minimize the chances of chronic disease development and death at an earlier stage due to obesity, the current trend is in the \textit{development of image-based automated self-diagnostic methods} for healthy weight monitoring. Specific interest is in the development of \textit{face-based non-intrusive} healthcare/ telemedicine solutions for smartphones~\cite{Mann2020,wen2013computational,bolukbacs2019bmi}. This interest has been spurred by the wide-scale integration of face recognition technology in smartphones, such as the iPhone X series~\cite{selfie}, for legitimate access to mobile users. 

Consequently, a number of studies~\cite{sidhpura2022face, siddiqui2022toward,adhikary2022bmi,pham2021bmi,yousaf2021estimation} have been proposed for gleaning BMI from facial images using machine and deep-learning models. 
For most of these studies, heavy-weight CNN models such as VGG, ResNet-50, InceptionNet, and XceptionNet have been used for BMI prediction from facial images. These studies suggested that BMI can be gleaned from facial images with an MAE in the range [$0.32$, $5.03$] across various facial-image datasets annotated with BMI information.
However, these aforementioned heavy-weight models require enormous space and computational complexity due to the millions of parameters and computations involved~\cite{10093861,9660033}. These requirements make their deployment on the resource-constrained mobile device challenging. There is a \textbf{need for a lightweight model} of compact size for facial-analysis-based BMI prediction to facilitate on-device deployment and weight monitoring using smartphone applications. 



\vspace{0.25 cm} \noindent\textbf{Our Contribution:}
To advance the research in the facial analysis-based self-diagnostic tools for maintaining a healthy weight, the contributions of this paper are as follows:
\begin{itemize}
\item Development of a lightweight ensemble of facial patch-based models (PatchBMI-Net) for BMI prediction that obtains the best trade-off between the performance, size, and inference time. 

\item Cross-comparison with the performance of SOTA heavy-weight models for BMI prediction from facial images in terms of performance, size, and inference time. To facilitate this, CNN models namely VGG16, ResNet50, Xception, and Efficient-NetB0 for BMI prediction are compared with our proposed PatchBMI-Net model in terms of performance, size, and inference time when deployed on a smartphone.
\item Experimental investigation of three publicly available datasets namely, VisualBMI~\cite{Kocabey}, IllinoisDOC~\cite{Illinois}, and FIW-BMI~\cite{JIANG2019183} consists of facial images annotated with BMI information. 
\end{itemize}

\section{Prior Work on BMI Inference from Facial Images}

Wen and Guo~\cite{Wen13} proposed an automatic facial image-based BMI prediction model by using geometry and ratio-based parameters derived from an Active Shape Model. These parameters included cheekbone-to-jaw width, width-to-upper face height ratio, perimeter-to-area ratio, and eye size. The proposed method was evaluated on the MORPH-II face dataset and obtained a Mean Absolute Error (MAE) in the range of $2.65$ to $4.29$. However, the BMI annotation for the MORPH-II dataset is not publicly available.

Jiang et al.~\cite{JIANG2019183} conducted a study comparing geometry-based features and deep learning-based models for BMI prediction. The authors used the FIW-BMI and Morph-II datasets and reported that deep-learning models consistently outperformed geometry-based features in predicting BMI. 

Kocabey et al.~\cite{Kocabey} proposed a facial-image-based BMI prediction model based on the Support Vector Regression trained on deep features extracted from the VGG model. Experimental investigations on the VisualBMI dataset suggested a Pearson correlation of $0.71$, $0.57$, and $0.65$ for males, females, and overall, respectively. 

Dantcheva et al.~\cite{8546159} designed an end-to-end Convolutional Neural Network (CNN) model for BMI prediction by modifying the ResNet architecture. Specifically, the last fully connected layer of the ResNet model was replaced with a single channel, and the loss function was changed to smooth L$_1$ loss for the regression task.  The VIP-attribute dataset was used to evaluate the model. The method obtained an MAE score of $2.32$, $2.30$, and $2.36$ for males, females, and overall, respectively.

Siddiqui et al.~\cite{siddiqui2020based,Siddiqui_2022_CVPR} evaluated various pre-trained models, such as VGG-19, ResNet, DenseNet, and  MobileNet for the deep feature extraction from the facial images for BMI prediction. The extracted deep features were used to train Support Vector Regression (SVR) and Ridge Regression (RR) models for BMI prediction on the Bollywood, VIP attribute, and VisualBMI datasets. The model obtained an average MAE in the range [$1.04$, $6.48$] across datasets. 
Sidhpura et al.~\cite{sidhpura2022face} used state-of-the-art pre-trained models such as Inception-v3, VGG-Face, VGG19, and Xception fine-tuned for BMI prediction. Illinois DOC, Arrest Records, and VIP attribute datasets were used for performance evaluation. The reported MAE for the IllinoisDOC, VIP attribute, and Arrest Records face datasets was in the range of [$2.82, 3.63$],  [$3.10, 3.91$], and [$3.73, 3.93$], respectively.

Yousaf et al.~\cite{yousaf2021estimation} predicted BMI using deep features obtained from facial regions such as the eyes, nose, lips, and brows. These facial regions were obtained using semantic segmentation based on a convolutional Neural Network (CNN). VGGFace and FaceNet-based CNN models were utilized to extract deep features for the facial regions. 
The deep features extracted from the different facial regions were pooled together using region-aware global pooling for the prediction. 
The results suggested that using region-aware global pooling over global average pooling enhanced the performance by $22\%$, $3\%$, and $63.09\%$ on VIP attributes, VisualBMI and Bollywood dataset, respectively, on BMI prediction. 

As can be seen, most of these aforementioned existing studies used heavyweight models (such as VGG, ResNet, and DenseNet) for BMI inference from facial images.

\begin{table}
\centering
\caption{Architecture of the custom Patch-based CNN model for BMI prediction for each facial patch.}
\scalebox{0.9}{
\begin{tabular}{lll}
\textbf{Layer}     & \textbf{Output Shape} & \textbf{Parameters}   \\ \hline
conv1(3,3)    & (32, 32, 32)  & 896\\                                                
max pooling(2,2) &  ( 32,16,16) & 0  \\
conv2(3,3)     &      ( 64,16,16) & 18496     \\   
max pooling(2,2) & (64,8,8)   & 0   \\     
Dropout(0.5)      &       (64,8,8) &           \\
Attention layer   &       (64,8,8) & 4096  \\
Flatten               &(4096) &        \\
Fully connected (linear)        &      (128)  & 524416         \\ 
Fully connected (linear)        &      (1)   &  128       \\ \hline

\textbf{Total}               &  5,48,032                        \\ \hline
\end{tabular}}
\label{tab:attention}
\end{table}

\begin{figure} [!t]
\begin{center}
\label{images}
\includegraphics[width=0.98\linewidth]{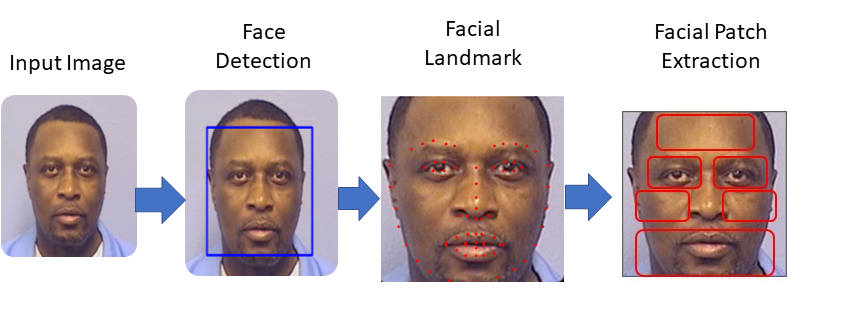}
\caption{Example face image with facial landmarks detection and facial patch extraction from forehead, left eye, right eye, left cheek, right cheek, and chin region.}
\label{figlandpatch}
\end{center}
\end{figure}

\begin{figure*}
    \centering
    \includegraphics[width=0.98\linewidth]{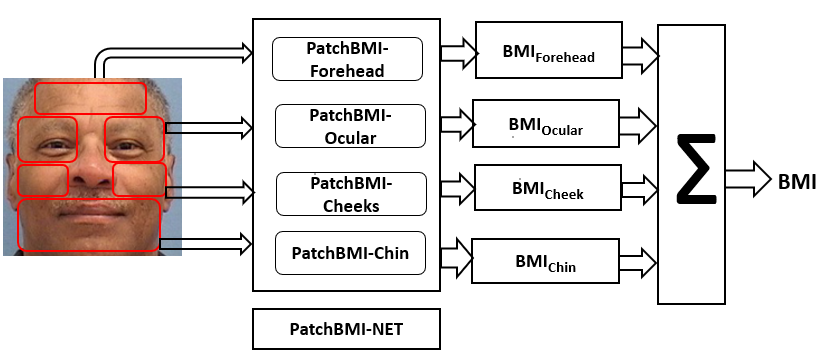}
    \caption{Overview of the steps involved in BMI prediction from a facial image.}
    \label{figoverview}
\end{figure*}

\section{Methodology: PatchBMI-Net}
In this section, we will discuss the steps involved in developing a proposed lightweight PatchBMI-Net model for facial analysis-based BMI prediction.

\subsection{Facial Patch Extraction}
\label{subsec:Facial Patch Extraction}

 Face detection and extraction of facial landmarks were done using the face-alignment method available in PyTorch package~\cite{Bulat_2017} that uses Haar features, defined as the variation in pixel intensity within the distinct rectangular region, along with the AdaBoost classifier for face detection. This is followed by the prediction of $68$ facial landmarks using a stacked hourglass network consisting of convolutional, residual, and up-sampling layers that output facial landmark coordinates in a given image. 

 After facial landmark coordinates prediction, the region of interest (ROI) was defined for the forehead, left and right eyes, left and right cheeks, and chin region based on the estimated coordinates for the facial patch extraction. The forehead ROI was defined as the area between landmarks $18$ and $25$. The left and right eye ROIs were defined as the area between landmarks $36$ and $39$ and $36$ and $41$, respectively.
 The chin ROI was defined as the area between landmarks $2$ and $8$. The left and right cheek ROIs were defined as the area between landmarks $2$ and $31$ and $14$ and $46$, respectively. Note, that we excluded the nose region because it was not contributing to BMI prediction as a part of the ensemble. Figure~\ref{figlandpatch} shows the subsequent facial patch extraction process. All the extracted facial patches from the forehead region, left and right eyes, left and right cheeks and chin region were used to train six lightweight convolutional neural network (CNN) models developed from scratch, explained next.

\subsection{Training Lightweight Convolutional Neural Networks~(CNNs)}
\label{subsec:Lightweight Model}
The architecture of the lightweight CNN model developed from scratch is shown in Table~\ref{tab:attention}. It consists of two convolutional layers followed by max pooling and the attention layer. The channel-wise attention mechanism used two $1\times1$ convolutions to dynamically weight feature channels in a layer. 
These layers are followed by two fully connected layers and the final output layer of $1$ channel for BMI prediction. This architecture is \textbf{determined based on empirical evidence}. This lightweight model was trained separately for each facial patch resized to $32\times32$ (region) i.e., six lightweight models were trained corresponding to forehead, left and right eyes, left and right cheeks, and chin region.  




\subsection{BMI Prediction}
At the testing stage, facial patches are extracted from each facial image as mentioned in subsections~\ref{subsec:Facial Patch Extraction}. The extracted patches are processed separately by the six lightweight models, trained for each facial patch independently, for BMI prediction. The final BMI is calculated by averaging these individual BMI predictions obtained from the ensemble of six lightweight models (pertaining to six facial patches), for each facial image. Figure~\ref{figoverview} shows the overview of the steps involved in BMI prediction from the facial images using our proposed lightweight ensemble. In this figure, PatchBMI-Forehead and PatchBMI-Chin are the individual lightweight models pre-trained for BMI prediction from the forehead and chin region, respectively. PatchBMI-Ocular and PatchBMI-Cheeks are trained separately for the left and right eye and cheek regions, respectively, in this work.

\section{Experimental Details}

\subsection{Datasets}


\begin{itemize}

\item \textbf{IllinoisDOC labeled face dataset~\cite{Illinois}:}  This publicly available facial image dataset consists of frontal and profile images of $68,149$ convicts. The dataset is annotated with gender, height (inches), weight (lbs), and date of birth. The database's mean BMI is $27.8$, and its standard deviation is $5.2$. Among all the subjects, $282$ were underweight ($16$ $<$ BMI $\geq$ $18.5$), $17,516$ were normal ($18.5$ $<$ BMI $\geq$ $25$), $24,166$ were overweight ($25$ $<$ BMI $\geq$ $30$), and $16,934$ were obese (BMI $>$ $30$). Figure~\ref{figdata} (first row) shows sample face images from this dataset.

\item \textbf{VisualBMI dataset~\cite{Kocabey}:} This dataset comprises of $4206$ face images with corresponding gender and BMI information collected from the web. Among all the subjects, seven were in the under-weight range ($16$ $<$ BMI $\geq$ $18.5$), $680$ were normal ($18.5$ $<$ BMI $\geq$ $25$), $1151$ were overweight ($25$ $<$ BMI $\geq$ $30$), $941$ were moderately obese ($30$ $<$ BMI $\geq$ $35$), $681$ were severely obese ($35$ $<$ BMI $\geq$ $40$) and $746$ were very severely obese ($40$ $<$ BMI). The subset of $2,896$ images was used as the training set and the rest of the $1302$ images were used as the test set. Training and testing subsets were balanced across gender. Figure~\ref{figdata} (second row) shows sample face images from this dataset.

\item \textbf{FIW-BMI Dataset~\cite{JIANG2019183}:} The FIW-BMI dataset consists of $7930$ facial images annotated with gender, height, and weight information. BMI values ranged widely from $18$ to $60$. The database's mean BMI is $30.8$, and its standard deviation is $6.97$. Among all the subjects, $43$ were underweight ($16$ $<$ BMI $\geq$ $18.5$), $1662$ were normal ($18.5$ $<$ BMI $\geq$ $25$), $2455$ were overweight ($25$ $<$ BMI $\geq$ $30$), and $3770$ were obese (BMI $>$ $30$). Figure~\ref{figdata} (third row) shows sample face images from this dataset.

The above publicly datasets are widely used for BMI prediction from facial images in existing studies.

\begin{figure} [!t]
\begin{center}
\label{images1}
\includegraphics[scale=0.90]{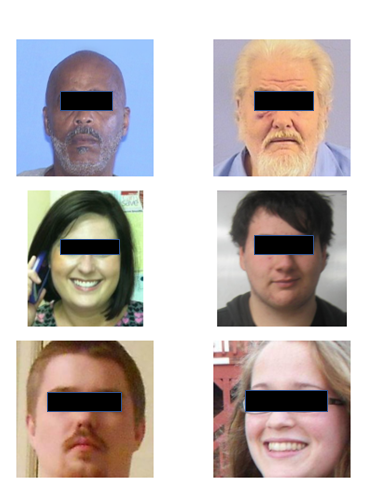}
\caption{Example of face images from  IllinoisDoc labeled face~\cite{Illinois},  VisualBMI~\cite{Kocabey}, and  FIW-BMI datasets~\cite{JIANG2019183} row-wise, respectively.}
\label{figdata}
\end{center}
\end{figure}


\end{itemize}

\subsection{Implementation Details}


For the development of the ensemble of the facial patch-based models (PatchBMI-Net), the facial images are resized to $224\times 224$ and converted to grayscale. Then the histogram equalization is applied to enhance the contrast of the image. 
This is followed by facial landmark detection and facial patch extraction from the forehead, left eye, right eye, jaw, left cheek, and right cheek, as discussed in section~\ref{subsec:Facial Patch Extraction}. The facial patches from each region were resized to $32\times32$ and input to the lightweight models trained separately for each facial patch, discussed in section~\ref{subsec:Lightweight Model}.  



Data transformations, including random horizontal flipping, and rotation, are applied to the training face images before facial patch extraction and lightweight ensemble training. The lightweight models are trained using the Mean squared Error (MSE) loss function, Adam optimizer, a learning rate of $0.001$, batch size of $32$, and the number of epochs were determined using the early stopping mechanism. 
These hyper-parameters were selected using a grid-search approach.

For cross-comparison, we also implemented and evaluated the performance of the baseline heavy-weight CNN models for BMI prediction. To this aim, VGG-16, ResNet-50, EfficientNet-B0, and Xception models pretrained on the ImageNet dataset are fine-tuned for BMI prediction. These models are fine-tuned by adding an adaptive average pooling layer followed by two fully connected layers of $512$ and $64$ channels, the dropout layer of $0.50$ based on empirical evidence, and the final output layer of size $1$ for BMI prediction. These models are trained using the Adam optimizer, batch size of $32$, a learning rate of $0.001$, a Mean Absolute Error (MAE) as the loss function, and the number of epochs determined using the early stopping mechanism.

\section{Experimental Validations}
The performance of the models for BMI prediction is evaluated using MAE (see eq.~\ref{eq:1}) computed as a measure of the difference between the BMI ($\hat{BMI}$) inferred by the model and the ground truth BMI averaged over $n$ face images in the test set. 

\begin{align}
MAE = \frac{\sum^{n}_{i=1}|\hat{BMI}_i-BMI_i|}{n}
\label{eq:1}
\end{align}

\subsection{Intra-dataset Evaluation}
Firstly, we evaluated the performance of our proposed PatchBMI-Net and heavy-weight deep-learning models in BMI inference from facial images. The mean absolute error (MAE) of VGG-16, EfficientNetB0, Xception, ResNet50, and our proposed PatchBMI-Net were compared in an intra-dataset setting (where training and testing subsets were from the same dataset).

On \textbf{visualBMI} dataset (see Table~\ref{mae_VBMI}), the VGG-16 model obtained an MAE of $6.94$, $6.39$, and $6.4$ for training, validation, and testing subsets, respectively, with a model size of $15.79$M. The EfficientNetB0 model demonstrated improved performance over all the models, obtaining an MAE of $3.39$, $4.81$, and $5.19$ for training, validation, and testing subsets, respectively, with a model size of $5.87$M. The Xception model demonstrated performance similar to EfficientNetB0 on the training dataset, with an MAE of $3.41$, but underperformed on the validation and testing subsets, with MAE values of $6.89$ each, with a larger model size of $23.46$M. The ResNet-50 model obtained an MAE of $3.51$, $5.5$, and $5.13$ for the training, validation, and testing datasets, respectively, with a model size of $26.16$M. Our proposed facial PatchBMI-Net, with a significantly smaller model size of $3.3$M, obtained MAE values of $6.45$, $6.47$, and $6.51$ for the training, validation, and testing subsets, respectively. 

 Similar study conducted by Siddiqui et al.~\cite{siddiqui2020based} on BMI prediction from facial images using pretrained models obtained average MAE of $5.87$ using ridge regression and support vector regression for BMI prediction using testing part of the VisualBMI dataset. The proposed PatchBMI-Net model obtained comparable performance with this existing study with a model size of about $5\times$ smaller.  

Thus, our proposed PatchBMI-Net model obtained a size reduction of about $5.4\times$ over all the heavyweight models with an average absolute decrease in MAE of only $0.79$ on the test set.



\begin{table}[t!]
\caption{Trade-off between MAE and size of proposed PatchBMI-Net over heavy-weight CNN models in BMI prediction on the visualBMI dataset.}
\begin{center}
\scalebox{0.9}{
\begin{tabular}{c|c|c|c|c}
\hline
\textbf{Model}        & \textbf{Training} & \textbf{Validation} & \textbf{Testing} & \textbf{Size} \\ \hline
VGG-16~\cite{Simonyan14c}     & 6.94  & 6.39 & 6.4    & 15.79M   \\
EfficientNetB0~\cite{tan2020efficientnet} & 3.39   & 4.81   & 5.19  & 5.87M  \\
Xception~\cite{chollet2017xception}   & 3.41  & 6.89     & 6.89   & 23.46M  \\
ResNet50~\cite{7780459}  & 3.51   & 5.5   & 5.13     & 26.16M\\ 
PatchBMI-Net (proposed)   & 6.45   & 6.47   & 6.51   & \textbf{3.3M} \\ 
Siddiqui et al.~\cite{siddiqui2020based}   &-   &-   &  5.87   &15.79M  \\ \hline
\end{tabular}}
\end{center}
\label{mae_VBMI}
\end{table}

On \textbf{IllinoisDOC} face dataset (see Table~\ref{mae_DOC}), the VGG-16 model obtained an MAE of $4.14$, $3.84$, and $3.85$ for training, validation, and testing subsets, respectively, with a model size of $15.79$M. The EfficientNet-B0 model exhibited improved performance, obtaining an MAE of $2.85$, $3.77$, and $3.76$ for training, validation, and testing subsets, respectively, with a model size of $5.87$M.
  The VGG-16 model obtained an MAE of $4.14$, $3.84$, and $3.85$ for training, validation, and testing subsets, respectively, with a model size of $15.79$M. The EfficientNetB0 model exhibited improved performance, obtaining an MAE of $2.85$, $3.77$, and $3.76$ for training, validation, and testing subsets, respectively, with a model size of $5.87$M. The Xception model demonstrated comparable performance with all the models for the validation and testing subsets, with MAE values of $3.75$ and $3.78$, respectively, with a larger model size of $23.46$M. The ResNet50 model obtained an MAE of $2.36$, $3.55$, and $3.56$ for the training, validation, and testing subsets, respectively, with a model size of $26.16$M. Lastly, our proposed facial PatchBMI-Net model, with a significantly smaller model size of $3.3$M, resulted in MAE values of $3.67$, $3.57$, and $3.58$ for the training, validation, and testing subsets, respectively. 

 Similar study conducted by Sidhpura et al.~\cite{sidhpura2022face} used state-of-the-art pre-trained models such as Inception-v3, VGG-Face, VGG19, and Xception fine-tuned for BMI prediction. They reported MAE for the IllinoisDOC datasets was about $3.63$ on the testing set. The proposed PatchBMI-Net model obtained similar performance with an absolute increase in MAE of $0.45$ on the test set.

 Thus, overall our proposed PatchBMI-Net model obtained on average a size reduction of about $5.4\times$ over all the heavyweight models along with the absolute average decrease in MAE of $0.16$ on the test set.
 

\begin{table}[t!]
\caption{Trade-off between MAE and size of the proposed facial PatchBMI-Net model over heavy-weight CNN models in BMI prediction on the IllinoisDOC face dataset.}
\begin{center}
\scalebox{0.9}{
\begin{tabular}{c|c|c|c|c}
\hline
\textbf{Model}        & \textbf{Training} & \textbf{Validation} & \textbf{Testing} & \textbf{Size} \\ \hline
VGG-16~\cite{Simonyan14c}   & 4.14   & 3.84   & 3.85  & 15.79M               \\
EfficientNetB0~\cite{tan2020efficientnet} & 2.85     & 3.77   & 3.76  & 5.87M                \\
Xception~\cite{chollet2017xception}       & 2.38     & 3.75     & 3.78   & 23.46M                  \\
ResNet50~\cite{7780459}     & 2.36     & 3.55  & 3.56   &26.16M              \\ 
PatchBMI-Net (proposed)   & 3.67   & 3.57   & \textbf{3.58}   & \textbf{3.3M} \\
Sidhpura et al.~\cite{sidhpura2022face}   &    & -   & 3.22   & - \\ \hline
\end{tabular}}
\end{center}
\label{mae_DOC}
\end{table}

\begin{table}[t!]
\caption{Trade-off between MAE and size of proposed facial PatchBMI-Net model over heavy-weight CNN models in BMI prediction on the FIW-BMI dataset.}
\begin{center}
\scalebox{0.9}{
\begin{tabular}{c|c|c|c|c}
\hline
\textbf{Model}        & \textbf{Training } & \textbf{Validation} & \textbf{Testing } & \textbf{Size} \\ \hline
VGG-16~\cite{Simonyan14c}     & 6.85    & 6.98     & 6.86    & 15.79M              \\
EfficientNetB0~\cite{tan2020efficientnet} & 3.19      & 4.44    & 4.48       & 5.87M            \\
Xception~\cite{chollet2017xception}       & 3.04      & 4.53       & 4.65    & 23.46M               \\
ResNet50~\cite{7780459}     & 3.06   & 4.81      & 4.83      &26.16M            \\ 
PatchBMI-Net (proposed)   & 6.14   & 5.76   & 5.98   & 3.3M \\ \hline
\end{tabular}}
\end{center}
\label{mae_FIW}
\end{table}

 On the \textbf{FIW-BMI} dataset, the VGG-16 model obtained an MAE of $6.85$, $6.98$, and $6.86$ for training, validation, and testing subsets, respectively, with a model size of $15.79$M. The EfficientNetB0 model exhibited improved performance, and obtained an MAE of $3.19$, $4.44$, and $4.48$ for training, validation, and testing datasets, respectively, with a model size of $5.87$M. The Xception model demonstrated better performance on the training dataset, with an MAE of $3.04$, but underperformed on the validation and testing subsets, with MAE values of $4.53$ and $4.65$, respectively, and a larger model size of $23.46$M. The ResNet50 model obtained an MAE of $3.06$, $4.81$, and $4.83$ for the training, validation, and testing datasets, respectively, with a model size of $26.16$M. Lastly, our proposed facial patch-based model, with a significantly smaller model size of $3.3$M, resulted in MAE values of $6.14$, $5.76$, and $5.98$ for the training, validation, and testing subsets, respectively. 

Overall, our proposed facial PatchBMI-Net model obtained an average reduction in the size of about $5.4\times$ along with an average absolute increase in MAE of $1.21$. 


In \textbf{summary}, our proposed PatchBMI-Net model obtained an average reduction in the size of $5.4\times$ with an average absolute increase in MAE of $0.087$ on the test set for all the datasets.

 \begin{table}[t!]
\caption{Cross-dataset evaluation of the proposed facial PatchBMI-Net model when trained on visualBMI and tested on IllinoisDOC and FIW-BMI datasets. Cross-comparison is done with the heavy-weight models trained for BMI prediction.}
\begin{center}
\scalebox{0.9}{
\begin{tabular}{c|c|c}
\hline
\textbf{Model}         & \textbf{Testing(Illinois)} & \textbf{Testing(FIWBMI)}  \\ \hline
VGG-16~\cite{Simonyan14c}      & 7.08 & 7.78     \\
EfficientNetB0~\cite{tan2020efficientnet}    & 4.30   & 6.87    \\
Xception~\cite{chollet2017xception}     & 5.24     & 6.43   \\
ResNet50~\cite{7780459}     & 4.41   & 5.24  \\ 
PatchBMI-Net (proposed)     & 5.97   & 6.62   \\  \hline
\end{tabular}}
\end{center}
\label{cross_VBMI}
\end{table}

\subsection{Cross Dataset Evaluation}
In this section, cross-dataset evaluation of the proposed facial PatchBMI-Net model was performed and compared with the performance of the heavy-weight models. 

As seen in Table~\ref{cross_DOC}, when trained on the IllinoisDOC dataset, the facial PatchBMI-Net model obtained an MAE of $6.61$ and $7.26$ when tested on FIW-BMI and VisualBMI datasets, respectively. 
In comparison with the heavy-weight models, our proposed model obtained an average decrease in the MAE of $4.38$ over cross-dataset evaluation.

As seen in Table~\ref{cross_VBMI}, when trained on the visualBMI dataset, the model obtained an MAE of $6.62$ and $5.97$ when tested on FIW-BMI and IllinoisDOC datasets, respectively. 
In comparison with the heavy-weight models, our proposed model obtained an increase in the MAE of only $0.36$ over cross-dataset evaluation.
 

Similarly, when our proposed model was trained on the FIW-BMI dataset (see Table~\ref{cross_FIWBMI}), it obtained an MAE of $6.27$ and $4.50$ when tested on the visualBMI and IllinoisDOC datasets, respectively. In comparison with the heavy-weight models, our proposed model obtained an average decrease in the MAE of $1.26$ on cross-dataset evaluation.


\begin{table}[t!]
\caption{Cross-dataset evaluation of the proposed facial PatchBMI-Net when trained on FIW-BMI and tested on IllinoisDOC and visualBMI datasets. Cross-comparison is done with the heavy-weight models trained for BMI prediction.}
\begin{center}
\scalebox{0.9}{
\begin{tabular}{c|c|c}
\hline
\textbf{Model}        & \textbf{Testing(Illinois)} & \textbf{Testing(VBMI)}  \\ \hline
VGG-16~\cite{Simonyan14c}      & 7.66 & 9.95     \\
EfficientNetB0~\cite{tan2020efficientnet}    & 4.35   & 7.76    \\
Xception~\cite{chollet2017xception}     & 6.81     & 6.90   \\
ResNet50~\cite{7780459}     & 4.31   & 5.44  \\ 
PatchBMI-Net (proposed)   & 4.50   & 6.27   \\  \hline
\end{tabular}}
\end{center}
\label{cross_FIWBMI}
\end{table}

\begin{table}[t!]
\caption{Cross-dataset evaluation of the proposed PatchBMI-Net when trained on IllinoisDOC and tested on visualBMI and FIW-BMI datasets. Cross-comparison is done with the heavy-weight models trained for BMI prediction.}

\begin{center}
\scalebox{0.9}{
\begin{tabular}{c|c|c}
\hline
\textbf{Model}         & \textbf{Testing(VBMI)} & \textbf{Testing(FIW-BMI)}  \\ \hline
VGG-16~\cite{Simonyan14c}       & 7.01 & 6.81     \\
EfficientNetB0~\cite{tan2020efficientnet}    & 9.83   & 9.45    \\
Xception~\cite{chollet2017xception}     & 16.89     & 28.12   \\
ResNet50~\cite{7780459}     & 6.38   & 6.06  \\ 
PatchBMI-Net (proposed)      & 7.26   & 6.61  \\  \hline
\end{tabular}}
\end{center}
\label{cross_DOC}
\end{table}

In \textbf{summary}, our proposed PatchBMI-Net demonstrated equivalent performance with heavy-weight models in the intra- and cross-dataset evaluation scenarios along with the size reduction of about $5.4\times$. 



\subsection{Inference Time Evaluation}
Finally, we also evaluated and compared the inference time of the PatchBMI-Net compared to other heavy-weight models evaluated in this study. For the purpose of this experiment, run-time performance is measured in terms of inference time in milliseconds (ms) for the baseline heavy-weight and proposed PatchBMI-Net on an Apple iPhone-$14$ smartphone with iOS version $16$. The models are deployed on the Apple iPhone-14 using PyTorch Mobile framework  \footnote{\url{https://pytorch.org/mobile/home/}} that provides an end-to-end workflow for training and deployment of machine learning models on edge devices.
Table~\ref{inference_time} shows the inference time of our proposed PatchBMI-Net in comparison to other heavy-weight models. The proposed PatchBMI-Net outperforms all other models in terms of faster inference time, requiring only $0.27$ms. This makes it roughly $2.2\times$ faster than the VGG-16, $3\times$ faster than EfficientNetB0, about $3\times$ faster than Xception, and approximately $3.7\times$ faster than ResNet50. On average, our proposed PatchBMI-Net is $3\times$ faster than other models in BMI prediction.

Thus, the proposed PatchBMI-Net offers a size and inference time reduction of about $5.4\times$ and $3\times$ over other heavy-weight models. This makes our proposed PatchBMI-Net model an attractive choice as a self-diagnostic AI-based weight monitoring tool for integration into smartphones.



\begin{table}[t!]
\caption{Evaluation of Inference time of the proposed PatchBMI-Net in comparison to the heavy-weight models when deployed on Apple iPhone-$14$ smartphone. The sizes of all the models are also listed for the sake of comparison. Our proposed model offers low latency over all the heavy-weight models used in this study.}
\begin{center}
\scalebox{0.9}{
\begin{tabular}{c|c|c}
\hline
\textbf{Model}        & \textbf{Inference time~(ms)} & \textbf{size}    \\ \hline
VGG-16~\cite{Simonyan14c}     & 0.6  & 15.79M    \\
EfficientNet-B0~\cite{tan2020efficientnet} & 1.2   & 5.87M     \\
Xception~\cite{chollet2017xception}   & 0.8  & 23.46M      \\
ResNet50~\cite{7780459}  & 1.0   & 26.16M  \\ 
PatchBMI-Net (Proposed)   & 0.27   & 3.3M       \\  \hline
\end{tabular}}
\end{center}
\label{inference_time}
\end{table}

\section{Conclusion and future work}

This research sought to develop a lightweight model for BMI predictions from facial images to facilitate on-device deployment and weight monitoring using resource-constrained smartphones. To this aim, we proposed a lightweight PatchBMI-Net model consisting of an ensemble of facial patch-based CNN models for BMI prediction. Thorough experiments in intra- and cross-dataset scenarios demonstrate that our proposed PatchBMI-Net can obtain equivalent performance with other heavy-weight models along with the size reduction of about $5.4\times$ and faster inference time of $3\times$. Thus, demonstrating the efficacy of the proposed model as a self-diagnostic tool for weight monitoring/ management using a smartphone application. As a part of future work, cross-comparison of the lightweight PatchBMI-Net model will be performed with the compact BMI prediction models obtained using existing neural network compression techniques, such as network pruning and knowledge distillation~\cite{neill2020overview}, to draw further insights. Further, the efficacy of our proposed lightweight model will be evaluated for other downstream image classification tasks. 

\small{
\flushend
\bibliographystyle{IEEE}
\bibliography{conference_101719}}
\end{document}